\documentclass[letterpaper, 10 pt, conference]{ieeeconf}  

\IEEEoverridecommandlockouts                              

\usepackage{amsmath} 
\usepackage{censor}
\usepackage{cleveref}
\usepackage{graphicx}
\usepackage{booktabs}

\usepackage{xcolor}

\title{\LARGE \bf
AnyCamVLA: Zero-Shot Camera Adaptation for Viewpoint Robust Vision-Language-Action Models
}

\author{Hyeongjun Heo$^{1}$, Seungyeon Woo$^{1}$, Sang Min Kim$^{1}$, Junho Kim$^{1}$,\\ Junho Lee$^{1}$, Yonghyeon Lee$^{2}$, and Young Min Kim$^{1}$
\thanks{This work was supported by Institute of Information \& communications Technology Planning \& Evaluation (IITP) grant funded by the Korea government(MSIT) [RS-2021-II211343, Artificial Intelligence Graduate School Program and RS-2025-25442338, AI star Fellowship Support Program (Seoul National University)]. Young Min Kim is the corresponding author.}
\thanks{$^{1}$H. Heo, S. Woo, S.M. Kim, J. Kim, J. Lee, and Y.M. Kim are with the Department of Electrical and Computer Engineering, Seoul National University, Seoul 08826, South Korea {\tt\footnotesize \{heo0224, yeon0857, tkdals9082, 82magnolia, twjhlee, youngmin.kim\}@snu.ac.kr}}%
\thanks{$^{2}$Y. Lee is with the Department of Mechanical Engineering, Massachusetts Institute of Technology, Cambridge, MA 02139, USA {\tt\footnotesize yhl@mit.edu}}%
}
\begin{document}

\maketitle
\thispagestyle{empty}
\pagestyle{empty}

\begin{abstract}

Despite remarkable progress in Vision-Language-Action models (VLAs) for robot manipulation, these large pre-trained models require fine-tuning to be deployed in specific environments. 
These fine-tuned models are highly sensitive to camera viewpoint changes that frequently occur in unstructured environments. 
In this paper, we propose a zero-shot camera adaptation framework without additional demonstration data, policy fine-tuning, or architectural modification. 
Our key idea is to virtually adjust test-time camera observations to match the training camera configuration in real-time. 
For that, we use a recent feed-forward novel view synthesis model which outputs high-quality target view images, handling both extrinsic and intrinsic parameters. 
This plug-and-play approach preserves the pre-trained capabilities of VLAs and applies to any RGB-based policy. 
Through extensive experiments on the LIBERO benchmark, our method consistently outperforms baselines that use data augmentation for policy fine-tuning or additional 3D-aware features for visual input.
We further validate that our approach constantly enhances viewpoint robustness in real-world robotic manipulation scenarios, including settings with varying camera extrinsics, intrinsics, and freely moving handheld cameras.

\end{abstract}


\section{Introduction}
Vision-Language-Action models (VLAs) trained on large-scale vision-language-action paired datasets have recently emerged as a promising paradigm for general embodied intelligence~\cite{brohan2022rt,zitkovich2023rt,team2024octo,kim2024openvla,black2024pi_0,intelligence2025pi_, bjorck2025gr00t}. 
VLAs leverage the vast scale of internet vision-language data to acquire rich perceptual and semantic priors, positioning VLAs as scalable foundation models for robotic control. 
Consequently, a growing paradigm is to adopt such pre-trained VLAs and fine-tune them for specific tasks and environments encountered by end users.
Instead of manually defining task-specific objectives or explicitly modeling 3D reconstruction, a small number of demo trajectories are sufficient to deploy the robot in real-world settings.

However, recent work shows that fine-tuned VLAs often overfit to the specific configurations provided in the training trajectory, instead of deducing the spatial context~\cite{xie2024decomposing,li2024evaluating,fei2025libero-plus,wang2025vlatest,gao2026taxonomy}.
We mainly focus on deviations of hardware arrangement between the camera and the robot, where minor shifts are unavoidable in everyday unstructured environments, such as homes or offices.
Even under minor variations, the performance degrades significantly. For example, we observed that a 3\,cm shift in the wrist camera can halve the success rate.
The inherent end-to-end nature of VLAs necessitates acquiring demo trajectories in the modified setup and fine-tuning the entire model even after small shifts.

To address this sensitivity, prior work trains policies with data collected from diverse viewpoints~\cite{khazatsky2024droid} or augments datasets using novel view synthesis~\cite{tian2024vista, chen2024roviaug}. However, improving viewpoint coverage through augmentation requires substantially more data and training, which is particularly costly for large VLAs. 
Another line of work improves robustness using geometry-aware or view-invariant representations, such as multi-view features, point clouds, or depth~\cite{seo2023multi,ze20243d,abouzeid2025geoaware,pang2025learning,wilcox2025adapt3r,yang2025fp3}. However, because VLAs build upon Vision-Language Models (VLMs) pre-trained on internet-scale RGB data, incorporating additional modalities often requires architectural modifications and limits the ability to fully leverage pre-trained priors~\cite{zhang2025vlas,zhang2025spatial}.

\begin{figure}[t!]
    \centering
    \includegraphics[width=1.0\linewidth]{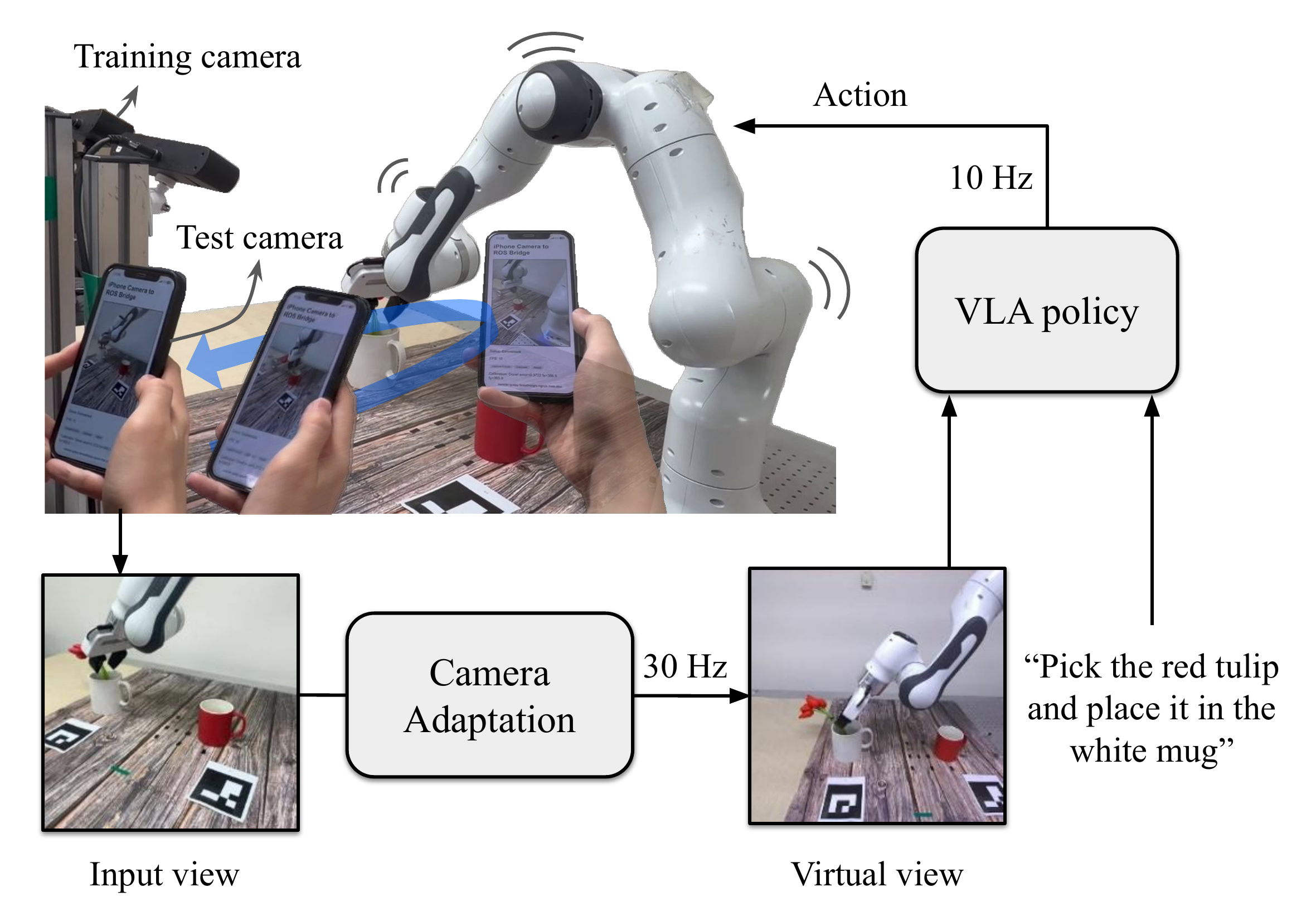}
    \vspace{-0.8cm}
    \caption{We present a zero-shot camera adaptation framework for VLAs that operates in real-time under various camera configuration changes from those used during training. At each control step, images from a test-time camera are synthesized into the training camera viewpoint at 30\,Hz, while the frozen VLA policy runs at 10\,Hz. Without any policy fine-tuning, our framework enables robust task execution across diverse setups—including different extrinsics, intrinsics, and freely moving handheld cameras such as an iPhone, ZED, and RealSense.}
    \vspace{-0.3cm}
    \label{fig:teaser}
\end{figure}
In this paper, we propose a simple yet effective framework for {\it zero-shot viewpoint adaptation} that requires no additional demonstrations, policy fine-tuning, or architectural modifications, as shown in~\Cref{fig:teaser}. 
The key idea is that we can freeze and leverage the latent space of pre-trained VLAs as long as we can virtually access images from the trained viewpoint.
While we can easily compensate for minor viewpoint changes with traditional image warping, inaccurate multi-view geometry can critically degrade overall performance.
Instead, we draw inspiration from recent progress in novel view synthesis in the computer vision community and generate precise projective warping of 3D geometry using a large view synthesis model~\cite{jin2025lvsm}.
The model is powerful at robustly handling various changes in pose, focal length, and lighting. 
Such a model handles only visual input, which is significantly smaller than the conventional VLAs. 
The viewpoint adaptation runs in real-time and can be plugged into the input of any VLAs to robustly process realistic variations.

Our approach naturally relieves the rigid requirements for camera-robot configuration without altering the pre-trained VLAs.
Remarkably, even under large viewpoint shifts (up to 15 cm translation and 60° rotation in camera extrinsics), the performance degradation of our method is minimal, showing the best task success rate over baselines on the LIBERO benchmark~\cite{liu2023libero}.
The view synthesis module is fast and delivers consistent performance across various camera configurations in real-world settings.
Together with the VLA module, everyday users can easily operate robots even with casual handheld capture.

\section{Related Work}

\subsection{Vision-Language-Action Models}
Vision-Language-Action models (VLAs) unify visual perception, language understanding, and robotic control by co-fine-tuning pre-trained Vision-Language Models (VLMs) on robotic trajectory data~\cite{brohan2022rt,zitkovich2023rt,team2024octo,kim2024openvla,black2024pi_0,intelligence2025pi_, bjorck2025gr00t}. 
By leveraging internet-scale vision-language pre-training, VLAs achieve impressive semantic generalization to novel objects, scenes, and natural language instructions without task-specific engineering.
However, VLAs exhibit significant fragility to distribution shifts, especially to camera viewpoint variations~\cite{fei2025libero-plus, wang2025vlatest, gao2026taxonomy}. 
As these models operate directly on 2D RGB observations without explicit spatial supervision or camera pose awareness, they tend to overfit to training viewpoints. 
Recent evaluations on LIBERO-Plus~\cite{fei2025libero-plus} and VLATest~\cite{wang2025vlatest} reveal substantial performance degradation under camera perturbations, with success rates dropping from over 90\% to below 30\%~\cite{fei2025libero-plus}. 
This camera viewpoint sensitivity poses a critical barrier to real-world deployment in unstructured environments where camera placements differ from trained environments.

\subsection{Viewpoint Robust Robotic Policies}
To address the viewpoint brittleness of visuomotor policies, recent work has explored two complementary directions: representation learning and data augmentation strategies.
Within the first direction, several methods enhance robustness through improved spatial representations beyond raw RGB images. 
Specifically, some works use explicit 3D representations such as point clouds~\cite{10610710, Ze2024DP3}, demonstrating substantial gains by converting the representation to the canonical reference frame which is consistent over viewpoints.
However, open-source large-scale robotic datasets lack point cloud data~\cite{o2024open}, limiting the scalability of point cloud-based VLAs. 
To overcome this reliance on depth sensors, GeoAwareVLA leverages implicit geometry-aware features from a pre-trained multi-view vision model~\cite{abouzeid2025geoaware}.
Parallel to these representational advancements, the second direction explores viewpoint generalization through data augmentation. 
Recent works~\cite{tian2024vista,chen2024roviaug} leverage novel view synthesis to augment training data across diverse viewpoints for viewpoint robust policy learning. However, training on augmented demonstrations remains costly for large-scale VLAs and still requires policy retraining.
Our work addresses these gaps by leveraging novel view synthesis as an input adaptation method that directly transforms observations at test time without requiring depth sensors, architectural modifications, or additional policy training.

\subsection{Novel View Synthesis for Robotics}
Novel view synthesis has been increasingly adopted in robotics for 3D scene reconstruction and understanding. Optimization-based methods such as Neural Radiance Fields (NeRF)~\cite{mildenhall2021nerf} and 3D Gaussian Splatting (3DGS)~\cite{10.1145/3592433} have demonstrated success in applications including grasping~\cite{IchnowskiAvigal2021DexNeRF, kerr2022evo, 10328050}, manipulation planning~\cite{shen2023F3RM, kim2025point2act}, and dynamic scene modeling~\cite{lu2024manigaussian, zhang2024dynamics}. However, these methods require per-scene training ranging from a few seconds to minutes, making them suitable primarily for offline applications where scene geometry can be pre-reconstructed before robot actions.
Recently, feed-forward novel view synthesis models have emerged as a scalable alternative to optimization-based approaches, directly predicting novel views or 3D representations from posed images in a single forward pass without per-scene optimization~\cite{charatan2024pixelsplat, chen2024mvsplat, xu2024depthsplat, smart2024splatt3r, jin2025lvsm}.
These advances enable real-time inference, making them well-suited for online robotic manipulation scenarios.

\section{Method}

\subsection{Problem Setup}
We consider a vision-language-action (VLA) policy $\pi_\theta$ trained with images captured from a set of cameras with known parameters. Let $\mathcal{C}^{\text{train}} = \{C_1^{\text{train}}, \ldots, C_M^{\text{train}}\}$ denote the camera parameters used during training, where each camera parameter $C_i^{\text{train}}$ includes both extrinsic parameters (pose) and intrinsic parameters (focal length, principal point, etc.). During training, the policy  $\pi_\theta(\cdot)$ parametrized by $\theta$ learns to map observations from these cameras to actions
\begin{equation}
a_t = \pi_\theta(I_t^{\text{train}}, l),
\end{equation}
where $I_t^{\text{train}} = \{I_{t,1}^{\text{train}}, \ldots, I_{t,M}^{\text{train}}\}$ are images captured by cameras with parameters $\mathcal{C}^{\text{train}}$ at time $t$, and $l$ is the language instruction.

At inference time, we encounter a different camera configuration $\mathcal{C}^{\text{test}} = \{C_1^{\text{test}}, \ldots, C_N^{\text{test}}\}$, where the camera parameters differ from those used during training. 
We assume that both $\mathcal{C}^{\text{train}}$ and $\mathcal{C}^{\text{test}}$ are known and expressed in the same coordinate frame, which is a reasonable assumption in practical deployment scenarios where camera calibration can be performed once during system setup.
Our goal is to enable the policy to perform robustly under this camera configuration change without retraining or fine-tuning, thereby preserving the pre-trained capabilities of the VLA across diverse tasks.

\subsection{Zero-Shot Camera Adaptation via Feed-Forward Novel View Synthesis}
To achieve our goal, we introduce a camera adaptation module $\mathcal{F}$ that transforms the test-time observations to match the training camera configuration:
\begin{equation}
\hat{I}_t^{\text{train}} = \mathcal{F}(I_t^{\text{test}}, \mathcal{C}^{\text{test}}, \mathcal{C}^{\text{train}}),
\end{equation}
where $\hat{I}_t^{\text{train}} = \{\hat{I}_{t,1}^{\text{train}}, \ldots, \hat{I}_{t,M}^{\text{train}}\}$ are the synthesized images as if they were captured from cameras with parameters $\mathcal{C}^{\text{train}}$. Importantly, the camera adaptation module $\mathcal{F}$ is flexible with respect to the number of input and output cameras, i.e., $N$ and $M$ can differ. The adapted policy inference then becomes:
\begin{equation}
a_t = \pi_\theta(\mathcal{F}(I_t^{\text{test}}, \mathcal{C}^{\text{test}}, \mathcal{C}^{\text{train}}), l).
\end{equation}
This formulation allows us to leverage the pre-trained policy without modification while adapting to arbitrary camera configurations at test time.

We implement the camera adaptation module $\mathcal{F}$ using a feed-forward novel view synthesis model~\cite{jin2025lvsm}, which can generate photorealistic images from arbitrary viewpoints given a set of source images and their corresponding camera parameters. 
Unlike optimization-based approaches such as Neural Radiance Fields (NeRF)~\cite{mildenhall2021nerf} that require per-scene optimization, a feed-forward model can synthesize novel views in a single forward pass, making it suitable for real-time robotic control.
As the model accepts a flexible number of input and output views with full camera parameters, it provides comprehensive coverage of the camera variation space encountered in real-world deployment. 

\Cref{fig:teaser} illustrates how our framework effectively works on different camera models.
At each time step during policy execution, our adaptation pipeline operates as follows: (1) capture input images $I_t^{\text{test}}$ from the current camera configuration, (2) virtually synthesize training-view images $\hat{I}_t^{\text{train}}$ using the camera adaptation module $\mathcal{F}$, (3) feed the synthesized images to the frozen policy to obtain action $a_t$, and (4) execute the action on the robot. Since the camera adaptation module operates at approximately 30\,FPS while typical VLAs run at around 10\,Hz~\cite{kim2024openvla, black2024pi_0, intelligence2025pi_}, our framework introduces negligible computational overhead and does not become a bottleneck in the control loop. Specifically, LVSM~\cite{jin2025lvsm} achieves a latency of 36.55\,ms for synthesizing 2 novel views from 2 input views at 256×256 resolution on an NVIDIA RTX 4090 GPU with BF16 mixed precision, corresponding to 27\,FPS.

Our zero-shot\footnote{We use "zero-shot" to denote that no new robot demonstrations or policy fine-tuning are required. The camera adaptation module for domain adaptation uses only multi-view images without robot demonstrations.} camera adaptation framework offers several key benefits. First, it requires no additional robot demonstrations or policy fine-tuning, reducing the computational cost and risk of catastrophic forgetting associated with retraining. Second, it operates in a plug-and-play manner, making it applicable to any RGB-based VLA without architectural modifications. Finally, unlike representation-centric methods that inject additional features, our approach preserves the original visual input modality, fully leveraging the pre-trained visual-linguistic reasoning capabilities of VLMs. 
\begin{figure}[t]
    \centering
    \includegraphics[width=1.0\linewidth]{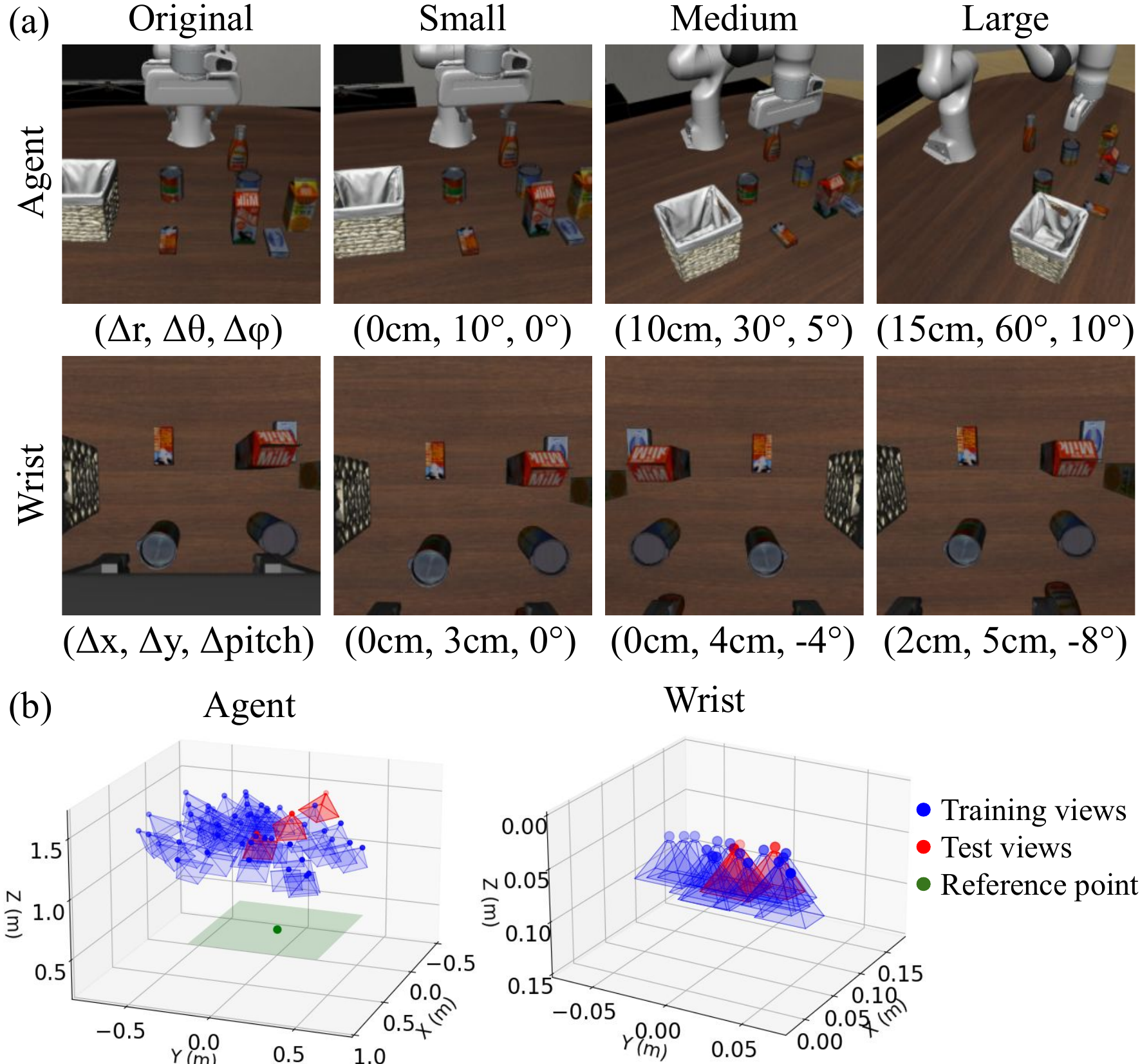}
    \caption{(a) Three perturbation levels on the agent and wrist camera. For the agent camera, we set the intersection point of the workspace surface and the camera $z$-axis as the center of spherical coordinates and apply perturbations to $(r, \theta, \phi)$ on camera poses. For the wrist camera, we perturb the $x$, $y$ coordinates and the pitch of the camera poses on the wrist camera frame. (b) Visualization of agent and wrist cameras in the viewpoint augmented LIBERO dataset. }
    \label{fig:experimental_setup}
    \vspace{-0.2cm}
\end{figure}
\begin{table*}[t]
\centering
\caption{Success rate (\%) on unseen agent camera viewpoint across LIBERO suites.}
\label{tab:agent_perturbation}
\resizebox{\textwidth}{!}{%
\begin{tabular}{l|ccc|c|ccc|c|ccc|c|ccc|c|ccc|c}
\toprule
& \multicolumn{4}{c|}{LIBERO-Spatial} & \multicolumn{4}{c|}{LIBERO-Object} & \multicolumn{4}{c|}{LIBERO-Goal} & \multicolumn{4}{c|}{LIBERO-Long} & \multicolumn{4}{c}{All Suites} \\
\cmidrule(lr){2-5} \cmidrule(lr){6-9} \cmidrule(lr){10-13} \cmidrule(lr){14-17} \cmidrule(lr){18-21}
Method & S & M & L & Avg & S & M & L & Avg & S & M & L & Avg & S & M & L & Avg & S & M & L & Avg \\
\midrule
OpenVLA-OFT & 98.2 & 92.4 & 91.0 & 93.9 & 78.6 & 42.4 & 15.4 & 45.5 & 84.4 & 75.8 & 68.0 & 76.1 & 79.6 & 8.8 & 10.4 & 32.9 & 85.2 & 54.9 & 46.2 & 62.1 \\
$\pi_{0.5}$ & 97.6 & 86.0 & 48.2 & 77.3 & 97.6 & 89.2 & 68.8 & 85.2 & 88.8 & 66.8 & 48.4 & 68.0 & 85.6 & 46.8 & 14.6 & 49.0 & 92.4 & 72.2 & 45.0 & 69.9 \\
$\pi_{0.5}^*$ & 93.6 & 95.4 & 94.2 & 94.4 & 94.0 & 95.2 & 94.0 & 94.4 & 86.4 & 86.0 & 85.0 & 85.8 & 76.4 & 75.0 & 71.6 & 74.3 & 87.6 & 87.9 & 86.2 & 87.2 \\
GeoAwareVLA & 76.6 & 84.2 & 78.2 & 79.7 & 97.4 & 94.6 & 93.4 & 95.1 & 88.4 & 86.2 & 85.2 & 86.6 & 86.0 & 83.6 & 79.0 & 82.9 & 87.1 & 87.2 & 84.0 & 86.1 \\
\midrule
Ours-OV & \textbf{99.0} & \textbf{98.4} & \textbf{97.4} & \textbf{98.3} & 95.8 & 63.4 & 49.6 & 69.6 & 96.2 & 94.0 & \textbf{96.4} & \textbf{95.5} & 89.6 & 79.4 & 67.6 & 78.9 & 95.2 & 83.8 & 77.8 & 85.6 \\
Ours-$\pi$ & 98.8 & 96.4 & 97.0 & 97.4 & \textbf{98.2} & \textbf{97.6} & \textbf{98.2} & \textbf{98.0} & \textbf{97.4} & \textbf{96.6} & 88.4 & 94.1 & \textbf{91.0} & \textbf{88.6} & \textbf{86.2} & \textbf{88.6} & \textbf{96.4} & \textbf{94.8} & \textbf{92.5} & \textbf{94.5} \\
\bottomrule
\end{tabular}
}
\vspace{-0.3cm}
\end{table*}
\section{Experiments}
\begin{table}[t]
\centering
\caption{Success rate (\%) on unseen wrist camera viewpoint on LIBERO-Long.}
\label{tab:wrist_perturbation}
\begin{tabular}{lcccc}
\toprule
Method & Small & Medium & Large & Average \\
\midrule
$\pi_{0.5}$ & 40.8 & 39.8 & 5.2 & 28.6 \\
$\pi_{0.5}^*$ & 84.0 & 84.0 & 81.2 & 83.1 \\
GeoAwareVLA & 1.6 & 5.0 & 9.0 & 5.2 \\
Ours-$\pi$ & \textbf{91.8} & \textbf{89.6} & \textbf{84.4} & \textbf{88.6} \\
\bottomrule
\end{tabular}
\vspace{-0.3cm}
\end{table}
In this section, we design experiments to answer the following four questions: (1) How much does feed-forward novel view synthesis help robustness on camera changes compared to baselines? (2) How does zero-shot adaptation framework outperform policy fine-tuning in terms of efficiency and flexibility? (3) Is feed-forward novel view synthesis the best approach for view adaptation? (4) Does our method perform effectively in real-world robot manipulation scenarios?

\subsection{Simulation Benchmark}
We evaluate on LIBERO benchmark~\cite{liu2023libero}, which is commonly used for VLAs. The LIBERO dataset consists of two input images: one from the agent camera facing the front of the robot arm, and one from the wrist camera mounted on the end effector. To assess the robustness of policies to viewpoint perturbations, we randomly perturb the camera poses at three levels.~\Cref{fig:experimental_setup} (a) illustrates example images under the three perturbation levels. During evaluation, we perturb either the agent camera (with wrist camera fixed) or the wrist camera (with agent camera fixed) to isolate the effect of viewpoint variations.
\subsubsection{Baselines}
\label{sec:baselines}
We compare our zero-shot camera adaptation framework with the following baselines:
\begin{itemize}
    \item \textbf{Base Policies.} We evaluate two VLAs as base policies: OpenVLA-OFT~\cite{kim2025fine} and $\pi_{0.5}$~\cite{intelligence2025pi_}. Both models are trained on the LIBERO dataset without camera perturbations.
    \item \textbf{Data Augmentation.} We fine-tune the pre-trained $\pi_{0.5}$ policy on viewpoint-augmented LIBERO data (denoted as $\pi_{0.5}^*$). For each robot action trajectory in the original LIBERO dataset, we render additional camera viewpoints to create a viewpoint-augmented training dataset. We randomly sample 50 agent camera viewpoints and 15 wrist camera viewpoints during fine-tuning for each camera perturbation. \Cref{fig:experimental_setup} (b) shows the camera distribution of each augmented dataset.
    \item \textbf{Representation-Centric Method.} We compare against GeoAware-VLA~\cite{abouzeid2025geoaware}, which replaces the RGB encoder with VGGT~\cite{wang2025vggt} to extract 3D-aware features from multi-view images. This approach incorporates implicit geometric reasoning for viewpoint invariance.
    \item \textbf{Ours.} We integrate LVSM~\cite{jin2025lvsm} decoder-only model as a visual adaptation module with both OpenVLA-OFT and $\pi_{0.5}$ (denoted as Ours-OV and Ours-$\pi$). 
\end{itemize}
To bridge the domain gap between LVSM's training data and synthetic LIBERO images, we fine-tune LVSM on custom multi-view simulation data while keeping the base policies frozen. We construct this multi-view dataset by rendering diverse camera viewpoints using objects and scene textures sourced from LIBERO-Plus~\cite{fei2025libero-plus}. Specifically, we replace all objects and surface textures (\textit{e.g.,} tabletop, background) in the original LIBERO scenes with assets from LIBERO-Plus while preserving the robot and workspace layout. Our custom dataset consists of 491 scenes, and for each scene, we randomly initialize the robot joint positions and render both agent and wrist cameras with 64 viewpoint variations. Importantly, this dataset contains only multi-view images without any robot action data, ensuring that LVSM learns purely geometric view synthesis without task-specific biases. Since none of these assets appear in the original LIBERO benchmark, our method operates on completely unseen objects and scenes at test time.

\subsubsection{Viewpoint Robustness Evaluation}

\Cref{tab:agent_perturbation} shows the success rate on agent camera perturbations across all LIBERO suites. Our methods consistently achieve the best performance across all suites and perturbation levels, demonstrating robust viewpoint generalization. While base policies experience drastic performance degradation as camera variation increases—for instance, OpenVLA-OFT drops from 85.2\% (small) to 46.2\% (large) and $\pi_{0.5}$ from 92.4\% to 45.0\%—our view adaptation method provides significant improvements for both base policies. Notably, Ours-$\pi$ achieves 94.5\% average success rate, while maintaining consistent performance across all perturbation levels. This demonstrates that our method effectively generalizes to any RGB-based VLA architecture.

To investigate the effectiveness of our method under different camera configurations, we also conduct wrist camera perturbations on LIBERO-Long, the most challenging suite with long-horizon tasks. As shown in \Cref{tab:wrist_perturbation}, our method achieves 88.6\% average success rate, outperforming both data augmentation and GeoAwareVLA. Interestingly, GeoAwareVLA shows a dramatic performance collapse under wrist camera perturbations, dropping to below 10\% across all perturbation levels despite achieving competitive performance (86.1\%) on agent camera perturbations in \Cref{tab:agent_perturbation}. 

We hypothesize this asymmetric behavior stems from the way VLAs learn to rely on different camera views. Comparing the base policy degradation between \Cref{tab:agent_perturbation} and \Cref{tab:wrist_perturbation} reveals that $\pi_{0.5}$ is significantly more sensitive to wrist camera changes (dropping to 28.6\%) than agent camera changes (dropping to 49.0\%), indicating a strong dependence on wrist camera features for action generation. This is intuitive given that the wrist camera provides close-range geometric and contact information critical for manipulation. GeoAwareVLA extracts 3D-aware features from multi-view images using VGGT, encoding geometric information such as relative camera transformations, depth, and multi-view correspondences. However, if the VLA learns to predominantly rely on wrist camera features during training as evidenced by the base policy behavior, the 3D geometric representations may become implicitly anchored to the wrist camera coordinate frame. In this case, when the agent camera viewpoint changes, the wrist-centric 3D representation remains valid, allowing GeoAwareVLA to function reasonably well. However, when the wrist camera is perturbed, the entire geometric reference frame becomes misaligned, causing the 3D-aware features to lose coherence and leading to complete policy failure. In contrast, our method synthesizes photorealistic RGB images in the canonical viewpoint, maintaining the visual appearance that the policy expects regardless of which camera is perturbed, thus avoiding coordinate frame dependencies entirely.

\subsubsection{Zero-Shot Adaptation vs. Policy Fine-Tuning}

To demonstrate the advantages of our zero-shot adaptation framework, we compare it against the fine-tuning approach used in data augmentation. We conduct experiments on LIBERO-Long, which consists of 10 tasks. Unlike the baseline data augmentation that uses viewpoint-augmented demonstrations from all 10 tasks, we evaluate fine-tuning $\pi_{0.5}$ with varying amounts of task coverage: 1, 5, and 10 tasks. For each configuration, we measure task success rates on novel viewpoints and the original training viewpoint throughout the fine-tuning process.
\begin{figure}[t!]
    \centering
    \includegraphics[width=1.0\linewidth]{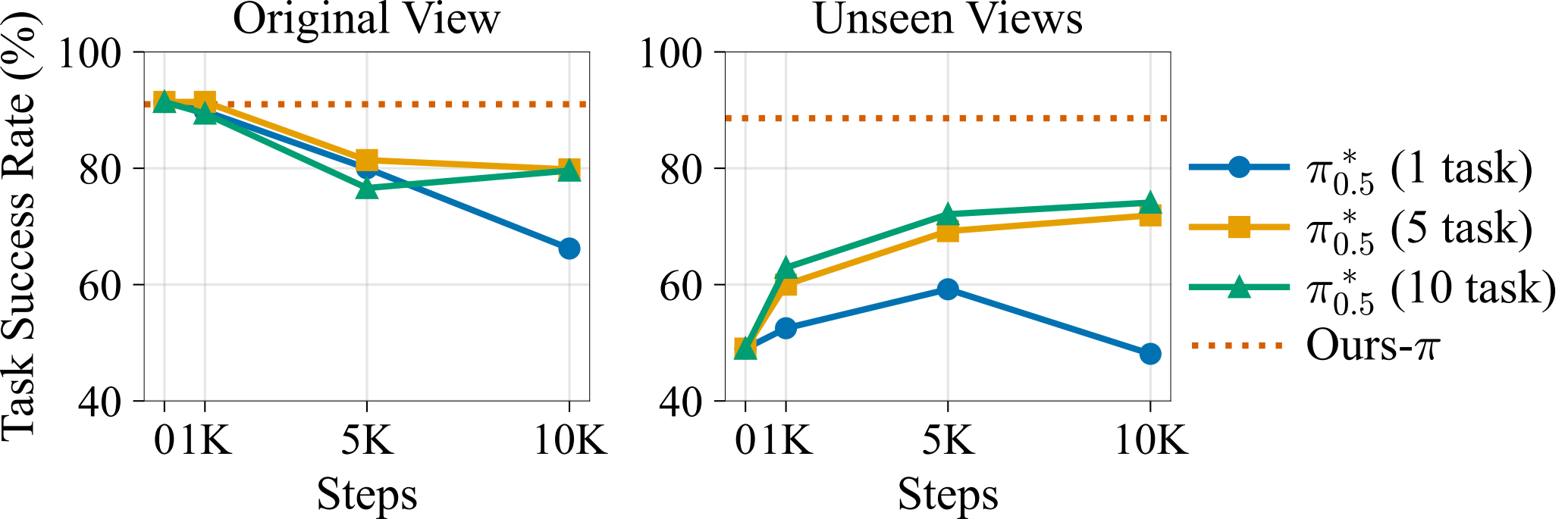}
    \vspace{-0.5cm}
    \caption{Success rate on LIBERO-Long with agent view variations by fine-tuning steps: (Left) Average task success rate on original view (Right) Average task success rate on all three unseen views.}
    \vspace{-0.3cm}
    \label{fig:finetuning}
\end{figure}

As shown in~\Cref{fig:finetuning}, fine-tuning with only 1 task actually decreases performance on novel viewpoints for the other 9 tasks. This reveals a critical limitation: viewpoint generalization learned from a single task does not transfer to other tasks, likely due to the task-specific visual patterns and object configurations. Furthermore, we observe a consistent decline in performance on the original viewpoint across all fine-tuning configurations as training progresses. This degradation indicates catastrophic forgetting~\cite{mccloskey1989catastrophic}, where the model gradually loses its original capabilities as it adapts to the augmented viewpoint distribution. Even when fine-tuning with all 10 tasks, the forgetting problem persists, suggesting that simply increasing the amount of viewpoint-augmented data cannot fully prevent the degradation of previously learned behaviors.

These results highlight two fundamental challenges with the fine-tuning approach: (1) achieving viewpoint generalization across diverse tasks requires demonstrations from a large number of tasks, making data collection expensive and impractical, and (2) the fine-tuning process inherently risks catastrophic forgetting, potentially degrading performance on original viewpoints. In contrast, our zero-shot adaptation framework circumvents both issues entirely. By keeping the policy frozen and adapting only the visual input, our method requires no additional robot demonstrations or training, eliminating data collection and computational training costs. 
Moreover, since the policy parameters remain unchanged, there is no risk of forgetting and the model retains its full capabilities on the original viewpoint while achieving robust generalization to novel viewpoints. This demonstrates that zero-shot visual adaptation is not only more efficient but also more reliable than fine-tuning VLAs for viewpoint robustness.

\begin{figure}[t!]
    \centering
    \includegraphics[width=1.0\linewidth]{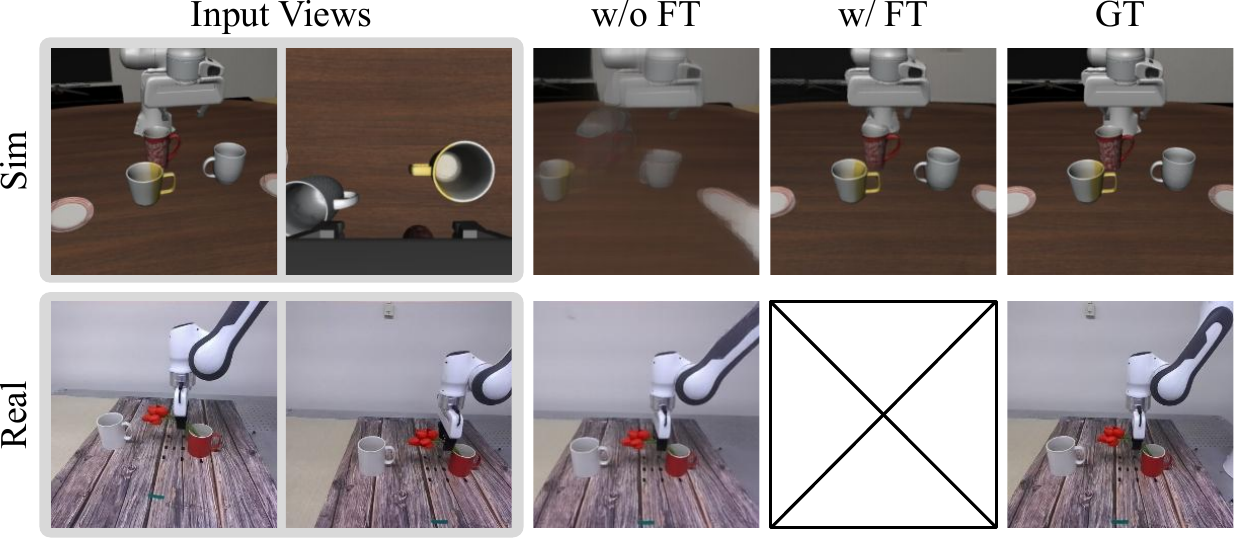}
    \vspace{-0.6cm}
    \caption{Qualitative results of LVSM on simulation and real images. Fine-tuning reduces the domain gap for input views on simulation.}
    \label{fig:ablation}
\end{figure}
\begin{table}[t]
\centering
\caption{Success rate (\%) and image quality on viewpoint adaptation ablation.}
\label{tab:ablation_view_adaptation}
\resizebox{\columnwidth}{!}{%
\begin{tabular}{lcccc|c}
\toprule
& \multicolumn{4}{c|}{Success Rate (\%)} & PSNR (dB)\\
Method & Small & Medium & Large & Average & Average \\
\midrule
$\pi_{0.5}$ (no adaptation) & 85.6 & 46.8 & 14.6 & 49.0 & 13.64 \\
Homography & 74.6 & 9.6 & 10.8 & 31.7 & 14.72 \\
Depth & 85.2 & 84.6 & 73.4 & 81.1 & 18.27 \\
Ours-$\pi$ (w/o FT) & 65.4 & 20.0 & 14.2 & 33.2 & 16.54 \\
Ours-$\pi$ & \textbf{91.0} & \textbf{88.6} & \textbf{86.2} & \textbf{88.6} & \textbf{23.20} \\
\bottomrule
\end{tabular}%
}
\vspace{-0.3cm}
\end{table}
\subsubsection{Ablation on Viewpoint Adaptation Methods}
We conduct an ablation study to compare our feed-forward novel view synthesis approach with alternative geometric methods for viewpoint adaptation. We evaluate three baseline approaches alongside our method on LIBERO-Long:
\begin{itemize}
    \item \textbf{Homography.} We assume the scene lies on a planar surface at the work table height and apply homography transformation to warp the input image to the target viewpoint.
    
    \item \textbf{Depth-Based Projection.} We leverage ground-truth depth maps available in simulation to construct a 3D point cloud by unprojecting RGB-D observations. The point cloud is then re-projected to the target viewpoint to synthesize the novel view.
    
    
    \item \textbf{Ours (w/o FT).} To isolate the effect of fine-tuning LVSM on our custom multi-view dataset, we evaluate the pre-trained LVSM model without domain adaptation. This ablation reveals the importance of bridging the domain gap between LVSM's training distribution and inputs from simulation.
    
    \item \textbf{Ours.} Our full method with LVSM fine-tuned on our custom multi-view dataset as described in~\Cref{sec:baselines}.
    \textit{Note that we only fine-tune to adapt to the synthetic benchmark dataset, and real-world deployment with the agent camera uses the original LVSM in a zero-shot manner.}
\end{itemize}
For both geometric baselines (homography and depth-based projection), unobserved regions in the synthesized views are filled using the Telea inpainting algorithm~\cite{telea2004image}. 

As shown in~\Cref{tab:ablation_view_adaptation}, our method significantly outperforms geometric baselines on image synthesis quality and task success rate across all perturbation levels. While depth-based projection shows considerable improvement over no adaptation and homography, the non-photorealistic artifacts from point cloud projection under large viewpoint shifts limit the visual understanding of the VLA, despite providing geometrically correct views. In contrast, feed-forward novel view synthesis leverages learned priors to accurately reconstruct 3D geometry from multi-view images while naturally synthesizing photorealistic images, which is supported by the superior image quality to baselines. This enables our method to maintain near-original performance ($88.6\%$ vs. $92.4\%$ on original view), demonstrating that VLAs can effectively process synthesized views when they are visually consistent with training data.

The critical role of fine-tuning LVSM is evident from the poor performance of the pre-trained LVSM without fine-tuning, as illustrated in \Cref{fig:ablation}. This degradation stems from two factors: (1) the domain gap between LVSM's training data (real-world images) and LIBERO simulation images (synthetic rendering with different lighting, textures, and materials), and (2) the distributional shift in camera configurations as LVSM is trained on continuous camera trajectories from indoor video sequences~\cite{zhou2018stereo}, whereas LIBERO uses a specific dual-camera configuration (agent and wrist) with distinct viewpoint characteristics. Without fine-tuning, LVSM fails to generate coherent views for the simulation domain, producing blurry or geometrically inconsistent images that confuse the policy. Fine-tuning on our custom multi-view dataset effectively bridges both gaps, enabling LVSM to adapt to the simulation domain while learning the specific geometric relationship between agent and wrist cameras, resulting in high-quality view synthesis and robust policy performance.

While our primary approach is zero-shot adaptation with frozen policies, we note that the one-time LVSM fine-tuning for domain adaptation is significantly more efficient than VLA fine-tuning. First, LVSM is substantially smaller (171M parameters) compared to VLAs (7B for OpenVLA-OFT, 3.3B for $\pi_{0.5}$), requiring less computational resources. More importantly, LVSM fine-tuning only requires multi-view images without any robot action labels, making data collection orders of magnitude easier and cheaper than collecting expert demonstrations for VLA training. This makes our approach highly practical: a single LVSM model fine-tuned once on multi-view data can serve as a universal view adapter for any downstream VLA.

\begin{figure}[t!]
    \centering
    \includegraphics[width=1.0\linewidth]{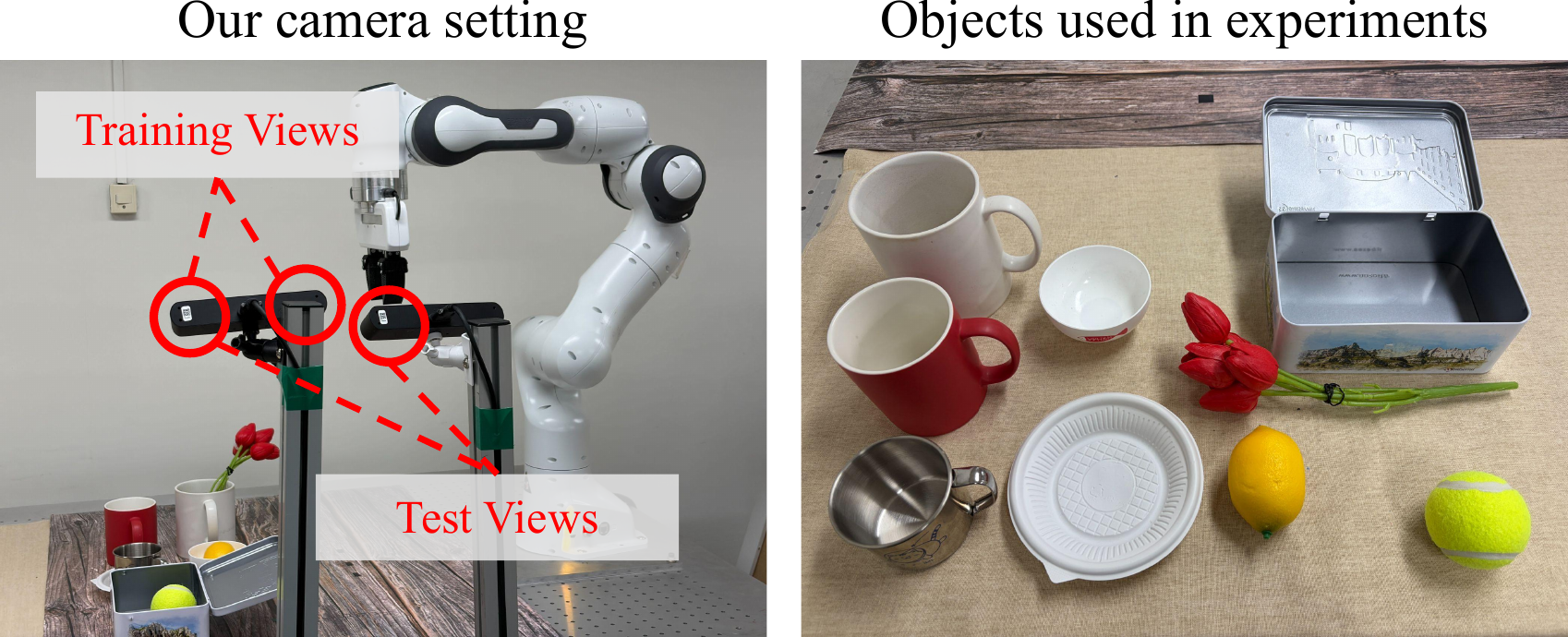}
    \caption{Experimental setup for real-world experiments. We use two input views for VLA fine-tuning. At test time, we change one input view.}
    \label{fig:real_setup}
\end{figure}
\begin{figure*}[t]
    \centering
    \includegraphics[width=0.9\linewidth]{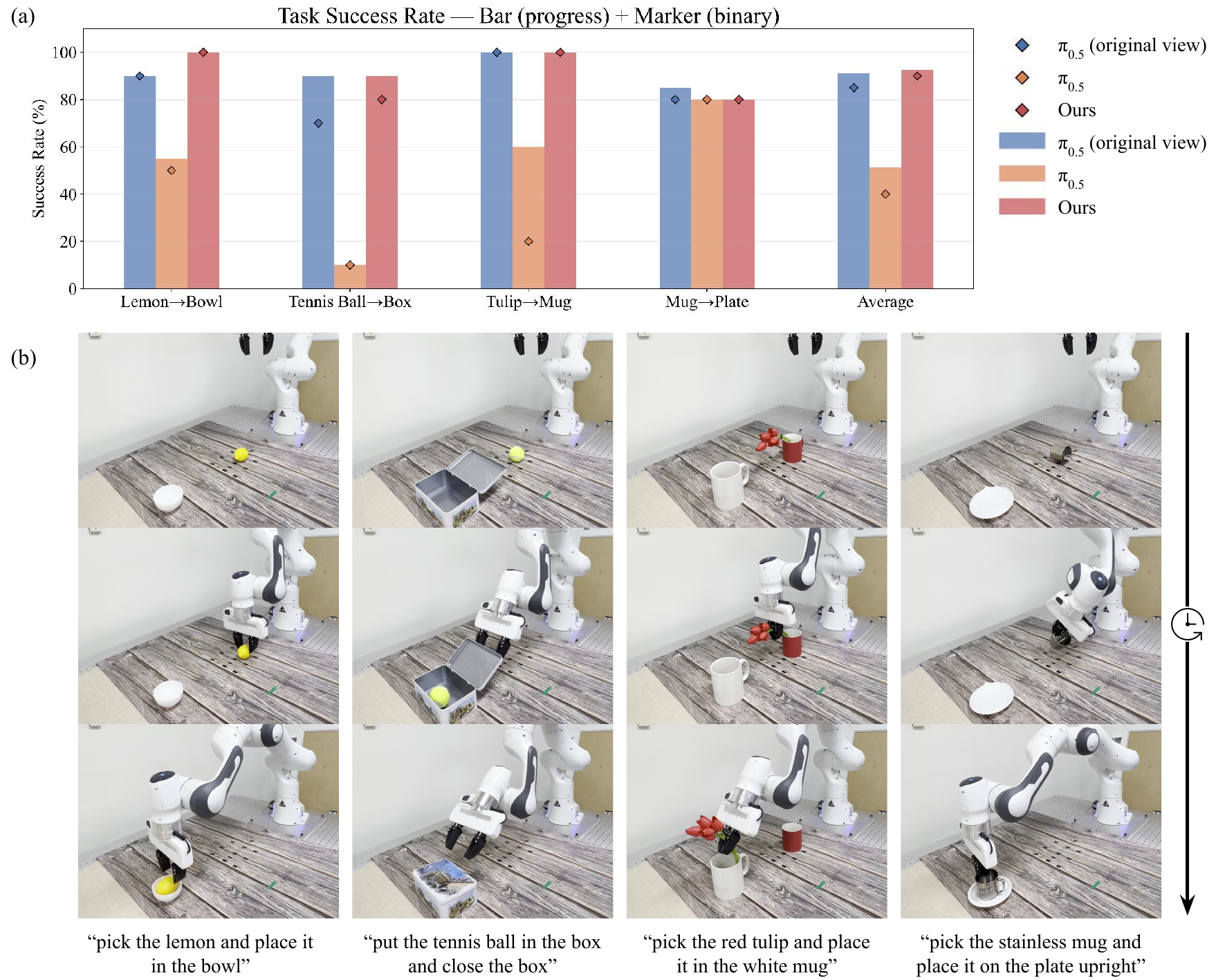}
    
    \caption{Results of performance improvement on novel viewpoint in real world. (a) Task success rate on novel viewpoints. Bar and marker each indicate progress and binary task success rate. (b) Real-world execution snapshots. Each column illustrates the sequential progression of a task from the initial state to the final state.}
    \label{fig:real_results}
\end{figure*}

\subsection{Real-World Experiments}
We conduct experiments in the real world to demonstrate that our zero-shot camera adaptation framework effectively improves the performance on viewpoint generalization, showing the practicability of our method in real-world manipulation scenarios.

\subsubsection{Experimental Setup}

We collect 20 demonstrations for each of the following tasks: (1) ``pick the lemon and place it in the bowl,'' (2) ``put the tennis ball in the box and close the box,'' (3) ``pick the red tulip and place it in the white mug,'' and (4) ``pick the stainless mug and place it on the plate upright.'' We use a Franka Panda robot arm with a Cartesian impedance controller from SERL~\cite{luo2024serl} for robot control, and an HTC Vive controller for teleoperation to collect the demo trajectories. During data collection, we fix a ZED2 stereo camera and use both views as input images. We fine-tune the $\pi_{0.5}$ base model using LoRA~\cite{hu2022lora} ($\sim$467M parameters) with our custom dataset via the AdamW optimizer~\cite{loshchilov2017decoupled} for 10,000 steps. At inference time, we replace the right-side view of the training camera with another ZED2 camera placed at a different position. Our experimental setup is shown in~\Cref{fig:real_setup}. 
Unlike the LIBERO benchmark, we adopt the original checkpoint of the LVSM decoder-only model trained on the RealEstate10K dataset~\cite{zhou2018stereo}.

\subsubsection{Performance Improvement on Novel Viewpoints}

We measure two metrics: \textit{Task Success Rate (Progress)}, which decomposes each task into multiple stages and assigns uniform partial credit per completed stage, and \textit{Task Success Rate (Binary)}, which considers a task successful only when the final state is achieved. For tasks (1), (3), and (4), we divide each task into two stages: (i) grasping the target object and (ii) placing it at the target location, assigning a score of 0.5 per stage. For task (2), we define three stages: (i) grasping the tennis ball, (ii) placing it inside the box, and (iii) closing the box, assigning a score of $\frac{1}{3}$ per stage.
Each policy is executed 10 times per task for evaluation.

\Cref{fig:real_results} presents the task success rates evaluated on novel viewpoints. 
For both metrics, the base model exhibits a consistent drop in task success rates across all tasks when evaluated on the novel viewpoint. Specifically, while the base model moves the end-effector to the vicinity of the target object or goal location, it fails to precisely grasp or place the object at the correct position, indicating that even a slight change in viewpoint is sufficient to degrade task performance. In contrast, our method maintains task success rates comparable to those achieved with the training viewpoint, demonstrating that our zero-shot camera adaptation framework effectively generalizes to novel viewpoints in real-world manipulation scenarios.

\subsubsection{Robustness on Various Handheld Cameras}
To further demonstrate the robustness of our method, we conduct an additional experiment where, unlike the fixed-camera setup in the previous evaluation, three different cameras (a ZED2 stereo camera, an Intel RealSense D435, and an iPhone 17 Pro) are held by hand and moved freely while the policy is being executed, as illustrated in~\Cref{fig:teaser}. The pose of each handheld camera is estimated using ArUco markers~\cite{garrido2014automatic} attached to the work table.  This experiment demonstrates that our method can successfully operate under dynamic camera conditions in real-time, generalizing not only to changes in camera extrinsics but also to variations in other camera parameters such as intrinsics and image characteristics across different camera models. Qualitative results of this experiment are provided in the supplementary video.
\section{Conclusion}

We present a zero-shot camera adaptation framework for VLAs 
under viewpoint variation from the training data. 
We employ feed-forward novel view synthesis models as the real-time camera-adaptation module while maintaining the policy inference loop.
Through extensive experiments, we demonstrate that our method enables viewpoint generalization without additional data collection, fine-tuning, or architectural modification of pre-trained VLAs.

Despite its effectiveness, our method has limitations. Our method can fail in cases when the reconstruction quality of novel view synthesis degrades, for instance, when the source view is limited to a single camera and the target viewpoint is far from the source view, or large occluded regions are present in the input images. Also, feed-forward novel view synthesis introduces a latency of approximately 30\,ms per frame, which may pose challenges in extremely dynamic scenarios, and requires additional GPU memory during inference. 
One open challenge lies in target viewpoint selection when the training camera configuration varies across demonstrations. Developing strategies for automatic target viewpoint selection is an important direction for future work.

\addtolength{\textheight}{-0.1cm}   


\bibliographystyle{IEEEtran}
\bibliography{references}

\end{document}